\pdfoutput=1
\documentclass[11pt]{article}

\usepackage{emnlp2021}

\usepackage{times}
\usepackage{latexsym}
\usepackage{times}
\usepackage{latexsym}
\usepackage{array} 
\usepackage{booktabs}
\usepackage{amsmath}
\usepackage{nccmath}
\usepackage{amssymb}
\usepackage[utf8]{inputenc}
\usepackage[T1]{fontenc}
\usepackage{graphicx}
\usepackage{multirow}
\usepackage{makecell}
\usepackage{pifont}
\usepackage{latexsym}
\usepackage{enumitem}
\usepackage{url}

\usepackage[T1]{fontenc}
\usepackage[utf8]{inputenc}

\usepackage{microtype}

\title{Time-aware Graph Neural Networks for Entity Alignment between Temporal Knowledge Graphs}

\author{Chengjin Xu \\
  University of Bonn / Germany\\
  \texttt{xuc@iai.uni-bonn.de} \\\And
  Fenglong Su \\
  National University of\\Defense Technology / China\\
  \texttt{sufenglong18@nudt.edu.cn} \\\And
Jens Lehmann \\
  University of Bonn\\
  Fraunhofer IAIS / Germany  \\
  \texttt{jens.lehmann}\\\texttt{@cs.uni-bonn.de} \\}

\begin{document}
\maketitle
\begin{abstract}
Entity alignment aims to identify equivalent entity pairs between different knowledge graphs (KGs). 
%A bunch of embedding-based approaches have been proposed for entity alignment, which embed entities and relations of different KGs into a vector space and measure the similarities between entity embeddings.
Recently, the availability of temporal KGs (TKGs) that contain time information created the need for reasoning over time in such TKGs. Existing embedding-based entity alignment approaches disregard time information that commonly exists in many large-scale KGs, leaving much room for improvement. In this paper, we focus on the task of aligning entity pairs between TKGs and propose a novel Time-aware Entity Alignment approach based on Graph Neural Networks (TEA-GNN). We embed entities, relations and timestamps of different KGs into a vector space and use GNNs to learn entity representations. To incorporate both relation and time information into the GNN structure of our model, we use a time-aware attention mechanism which assigns different weights to different nodes with orthogonal transformation matrices computed from embeddings of the relevant relations and timestamps in a neighborhood. Experimental results on multiple real-world TKG datasets show that our method significantly outperforms the state-of-the-art methods due to the inclusion of time information. Our datasets and source code are available at \url{https://github.com/soledad921/TEA-GNN}
\end{abstract}

\section{Introduction}~\label{intro}
Knowledge Graphs (KGs) provide a means for structured knowledge representation through connected nodes via edges. The nodes represent entities and the edges connecting these nodes denote relations. A KG stores facts as triples of the form $(e_s,r,e_o)$, where $e_s$ is the subject entity, $e_o$ is the object entity, and $r$ is the relation between entities. 
Many large-scale KGs including YAGO~\cite{YAGO} and DBpedia~\cite{Dbpedia} have been established and are widely used in NLP applications, e.g., question answering and language modeling~\cite{Review}. 
%and FreeBase~\cite{bollacker2008freebase}.

Since most KGs are developed independently and many of them are supplementary in contents, one of core challenges of KGs is to align equivalent entity pairs between different KGs. To address this issue, embedding-based approaches are leveraged to model entities and relations across multiple KGs and measure the similarities between entities~\cite{OpenEA}. It has been proven that the utility of  multi-relation information is helpful for an effective entity alignment approach~\cite{MRAEA}. 

\begin{figure}[t]
\centering
\includegraphics[width=7.7cm]{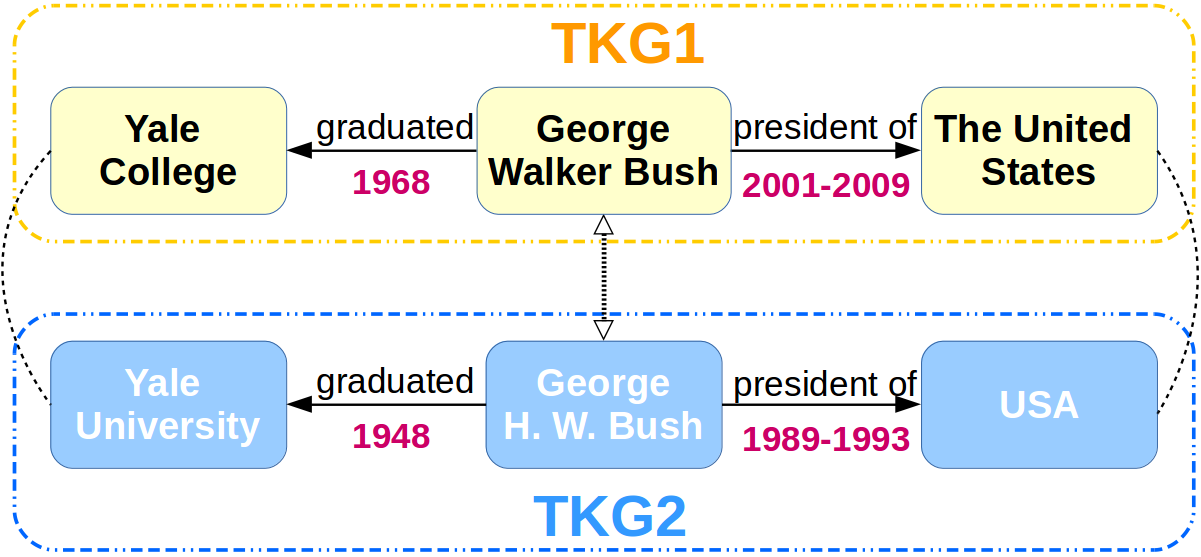} 
\caption{Illustration of the limitation of the existing time-agnostic entity align approaches.}
\label{fig:EAinTKG}
\end{figure}

In addition to relation information, many KGs including YAGO3~\cite{YAGO3}, Wikidata~\cite{Wikidata} and ICEWS~\cite{ICEWS} also contain time information between entities, i.e., some edges between entities have two properties, relation and time as shown in Figure~\ref{fig:EAinTKG}. Facts in such temporal KGs (TKGs) can be represented as quadruples shaped like $(e_s,r,e_o,\tau)$ where $\tau$ denotes the timestamp. Noteworthily, timestamps in most TKGs are presented in Arabic numerals and have similar formats. Thus, timestamps representing the same dates across multiple TKGs can be easily aligned by manually uniforming their formats. 
%That is to say, timestamps in different TKGs are naturally aligned. 

However, the existing embedding-based entity alignment approaches disregard time information in TKGs, leaving much room for improvement. Taking the case in Figure~\ref{fig:EAinTKG} as an example, given two entities, \text{George H. W. Bush} and \text{George Walker Bush}, existing in two TKGs respectively, time-agnostic embedding-based approaches are likely to ignore time information and wrongly recognize these two entities as the same person in the real world due to the homogeneity of their neighborhood information.

To address this issue, an intuitive solution is to incorporate time information into entity alignment models. Inspired by the recent successful applications of GNN models in entity alignment, in this paper, we propose a novel Time-aware Entity Alignment approach based on Graph Neural Networks (TEA-GNN) for entity alignment between TKGs. Different from some temporal GNN models which discretize temporal graphs into multiple snapshots, we treat timestamps as properties of links between entities. We first map all entities, relations and timestamps in TKGs into an embedding space. To incorporate relation and time information into the GNN structure, we utilize a time-aware attention mechanism which assigns different importance weights to different nodes within a neighborhood according to orthogonal transformation matrices computed with the embedddings of the corresponding relations and timestamps. To further integrate time information into the final entity representations, we concatenate output features of entities with the summation of their neighboring time embeddings to get multi-view entity representations.

Specifically, we create a reverse relation $r^{-1}$ for each relation $r$ to integrate direction information. And a time-aware fact involving a time interval $(e_s,r,e_o,[\tau_{b},\tau_e])$, where $\tau_{b}$ and $\tau_e$ denote the begin and end time, is separated into two quadruples $(e_s,r,e_o,\tau_{b})$ and $(e_o,r^{-1},e_s,\tau_{e})$, which represent the begin and the end of the relation, respectively. In this way, TEA-GNN can adapt well to datasets where timestamps are represented in various forms: time points, begin or end time, time intervals.

To verify our proposed approach, we evaluate TEA-GNN and its time-agnostic variant as well as several state-of-the-art entity alignment approaches on real-world datasets extracted from ICEWS, YAGO3 and Wikidata. Experimental results show that TEA-GNN significantly outperforms all baseline models with the inclusion of time information. To the best of our knowledge, this work is the first attempt to perform entity alignment between TKGs using a time-aware embedding-based approach.

\section{Related Work}
\subsection{Knowledge Graph Embedding}
KG embedding (KGE) aims to embed entities and relations into a low-dimensional vector space and measure the plausibility of each triples $(e_s,r,e_o)$ by defining a score function. A typical KGE model is TransE~\cite{TransE} which is based on the assumption of $e_s+r\approx e_o$. In addition to translational KGE models including TransE and its variants~\cite{TransH,TransComplex,SpacE}, other KGE models can be classified into semantic matching models~\cite{DISTMULT,GeomE} or  neural network-based models~\cite{ConvE,RGCN}.

With the development of TKGs, TKG embedding (TKGE) draws increasing attention~\cite{leblay,ATiSE,TERO,TeLM,TComplEx}. An example of typical TKGE models is TTransE~\cite{leblay} which represents timestamps as latent vectors with entities and relations and incorporates time embeddings into its score function $||e_s+r+\tau-e_o||$. The success of TKGE models shows that the inclusion of time information is helpful for reasoning over TKGs.
\subsection{Graph Neural Network}
Benefitting from the ability to model non-Euclidean space,
GNN has become increasingly popular in many areas, including social networks and KGs~\cite{RGCN}. Graph Convolutional Network (GCN)~\cite{GCN} is an
extension of GNN, which generates node-level embeddings
by aggregating information from the nodes’ neighborhoods. Furthermore, Graph Attention Network (GAT)~\cite{GAT} employs a self-attention mechanism to calculate the hidden representations of each entity by attending over its neighbors.

With the success of these GNN models in the
static setting, we approach further practical scenarios where
the graph temporally evolves. Existing approaches~\cite{GCN-LSTM, DGCN,EvolveGCN,TeMP} generally discretize a temporal graph into multiple static snapshots in a timeline and utilize a combination of GNNs and recurrent architectures (e.g., LSTM),
whereby the former digest graph information and the latter handle dynamism.

\subsection{Knowledge Graph Alignment}
Many KG alignment approaches are proposed to find equivalent entities across multiple KGs by measuring the similarities between entity embeddings. Most embedding-based entity alignment approaches can be classified into two categories, i.e., translational models and GNN-based models.

Typical translational entity align models are based on embeddings learned from TransE and its variants. MTransE~\cite{MTransE} learns a mapping between two separate KGE spaces. JAPE~\cite{JAPE} proposes to jointly learn structure embeddings and attribute embeddings in a uniform optimization objective. IPTransE~\cite{IPTransE} and BootEA~\cite{BootEA} employ a semi-supervised learning strategy which iteratively label new entity alignment as supervision. The main limitation of translational models is their inability of modeling 1-n, n-1 and n-n relations. Besides, they may lack to exploit the global view of entities since TransE is trained on individual triples.

Many recent studies introduce GNNs into entity alignment task, which is originated with the ability to model global information of graphs. GCN-Align utilizes GCNs to embed entities of each KG into a unified vector space without the prior knowledge of relations. After that, a bunch of GCN-based approaches are proposed to incorporate relation information into GCNs. HGCN~\cite{HGCN} jointly
learn both entity and relation representations via a GCN-based framework and RDGCN~\cite{RDGCN} construct a dual relation graph for embedding learning. MuGNN~\cite{MuGNN}, MRAEA and RREA~\cite{MRAEA,RREA} assign different weight coefficients to entities according to relation types between them, which empowers the models to distinguish the importance between different entities. Our framework TEA-GNN adopts a similar idea with additional time information.

\section{Problem Formulation}

Formally, a TKG is represented as $\mathcal{G}=(\mathcal{E},\mathcal{R},\mathcal{T},\mathcal{Q})$ where $\mathcal{E}$, $\mathcal{R}$ and $\mathcal{T}$ are the sets of entities, relations and timestamps, respectively. $\mathcal{Q}\subset \mathcal{E}\times\mathcal{R}\times\mathcal{E}\times\mathcal{T}$ is the set of factual quadruples. Let $\mathcal{G}_1=(\mathcal{E}_1,\mathcal{R}_1,\mathcal{T}_1,\mathcal{Q}_1)$ and $\mathcal{G}_2=(\mathcal{E}_2,\mathcal{R}_2,\mathcal{T}_2,\mathcal{Q}_2)$ be two TKGs, and $\mathcal{S}=\{(e_{i1}, e_{i2})|e_{i1}\in\mathcal{E}_1,e_{i2}\in\mathcal{E}_2\}$ be the set of pre-aligned entity pairs between $\mathcal{G}_1$ and $\mathcal{G}_2$. As mentioned in Section~\ref{intro}, timestamps in different TKGs can be easily aligned by manually uniforming their formats. A uniform time set $\mathcal{T}^*=\mathcal{T}_1\cup\mathcal{T}_2$ can be constructed for both TKGs. Therefore, two TKGs can be renewed as $\mathcal{G}_1=(\mathcal{E}_1,\mathcal{R}_1,\mathcal{T}^*,\mathcal{Q}_1)$ and $\mathcal{G}_2=(\mathcal{E}_2,\mathcal{R}_2,\mathcal{T}^*,\mathcal{Q}_2)$ sharing the same set of timestamps. The task of time-aware entity alignment aims to find new aligned entity pairs between $\mathcal{G}_1$ and $\mathcal{G}_2$ based on the prior knowledge of $\mathcal{S}$ and $\mathcal{T}^*$.

\section{The Proposed Approach}
To exploit both relation and time information for entity alignment, we first create a reverse link for each link so that each pair of reverse links between entities can represent relation directions and handle the begin and end of the relation. An orthogonal transformation-based time-aware attention mechanism is employed in each GCN layer to assign different weights to entities according to relation and time information between them. Finally, entity alignments are predicted by applying a distance function to multi-view representations of entities.
            \begin{figure*}[t!]
\centering
\includegraphics[width=15.5cm]{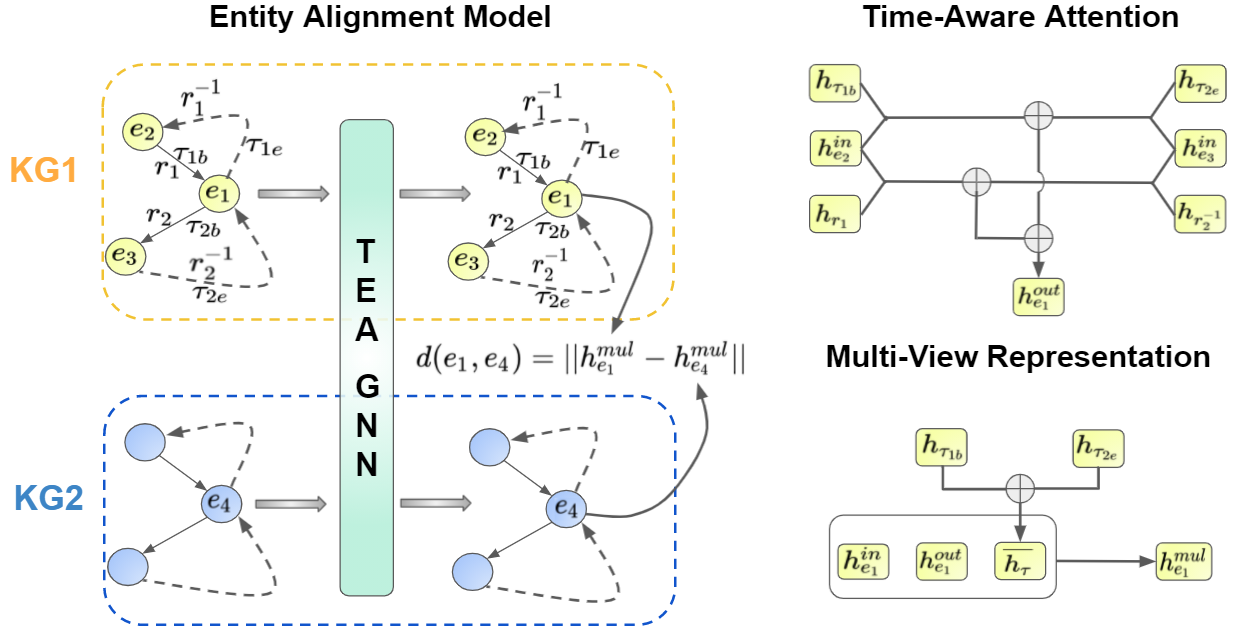} 
\caption{Framework of our approach. Dashed arrows represent the created reverse links.}
\label{fig:framework}
\end{figure*}

\subsection{Reverse Link Generation}
Time information $\tau$ in a temporal fact $(e_s, r, e_o, \tau)$ can be represented in various forms, e.g., time points, begin or end time and time intervals. A time interval is shaped like $[\tau_b,\tau_e]$ where $\tau_b$ and $\tau_e$ denote the actual begin time and end time of the fact, respectively. A time point can be represented as $[\tau_b,\tau_e]$ where $\tau_b=\tau_e$. Noteworthily, we represent a begin or end time as $[\tau_b,\tau_0]$ or $[\tau_0,\tau_e]$ where $\tau_0\in\mathcal{T}^*$ is the first time step in the time set denoting the unknown time information. A fact without known time information can be denoted as $(e_s, r, e_o, [\tau_0,\tau_0])$ to deal with heterogeneous temporal knowledge bases where a significant amount of relations might be non-temporal. 

In order to integrate relation direction, we create a reverse relation $r^{-1}$ for each relation $r$ and extend the relation set $\mathcal{R}=\{r_0,r_1,\cdots,r_{|\mathcal{R}|-1}\}\rightarrow\{r_0,r_0^{-1},\cdots,r_{|\mathcal{R}|-1},r^{-1}_{|\mathcal{R}|-1}\}$. And each fact $(e_s, r, e_o, [\tau_b,\tau_e])$ is decomposed into two quadruples $(e_s,r,e_o,\tau_b)$ and $(e_o,r^{-1},e_s,\tau_e)$ to handle the begin and the end of the relation, respectively. 

\subsection{Time-Aware Attention Network}
We map all of entities, relations (including reverse relations) and time steps in both TKGs into a same vector space $\mathbb{R}^{k}$ where $k$ denotes the embedding dimension. Embeddings of the entity $e_i$, relation $r_j$, time step $\tau_m$ are denoted as $h_{e_i},h_{r_j},h_{\tau_m}\in\mathbb{R}^k$.

Some recent studies~\cite{Orthogonal,OTEA} show
that orthogonal transformation matrix is desirable and robust when transforming one isomorphic embedding to another. Thus, for each relation embedding $h_{r_j}$ and time embedding $h_{\tau_m}$, we define the corresponding orthogonal transformation matrices $M_{r_j},M_{\tau_m}\in\mathbb{R}^{k\times k}$ as follows,
\begin{equation}
\begin{aligned}
M_{r_j} = I-2h_{r_j}h_{r_j}^{T}, \ M_{\tau_m} = I-2h_{\tau_m}h_{\tau_m}^{T},
\end{aligned}
\end{equation}
where embeddings $h_{r_j}$ and $h_{\tau_m}$ are normalized to ensure $h_{r_j}^{T}h_{r_j}=h_{\tau_m}^{T}h_{\tau_m}=1$. By doing this, we can easily prove that transformation matrices $M_{r_j}$ and $M_{\tau_m}$ are orthogonal. Taking $M_{\tau_m}$ as an example, we can obtain
\begin{align}
&M_{\tau_m}^{T}M_{\tau_m} = (I-2h_{\tau_m}h_{\tau_m}^{T})^{T}(I-2h_{\tau_m}h_{\tau_m}^{T})\nonumber\\
&=I-4h_{\tau_m}h_{\tau_m}^{T}+4h_{\tau_m}h_{\tau_m}^{T}h_{\tau_m}h_{\tau_m}^{T}=I.
\end{align}
By using such orthogonal transformation matrices, the norms and the relative distances of entities can remain unchanged after transformation, i.e.,
\begin{equation}
\begin{aligned}
&||h_{e_i}M_{\tau_m}|| = ||h_{e_i}||,\\
&h_{e_i}^{T}h_{e_j} = (h_{e_i}M_{\tau_m})^{T}(h_{e_j}M_{\tau_m}).
\end{aligned}
\end{equation}

A time-aware attention mechanism is used to integrate both time and relation information into entity representations by assigning different weights to different neighboring nodes according to the orthogonal transformation matrices of relations and timestamps of the corresponding inward links. 
In the case of Figure~\ref{fig:framework}, the inward links in the neighborhood of the entity $e_1$ include $(e_2,r_1,e_1,\tau_{1b})$ and $(e_3,r_2^{-1},e_1,\tau_{2e})$ in which $e_1$ performs as the object entity. We define the time-specific weighted importance $\alpha_{i,j,m}$ and the relation-specific weighted importance $\beta_{i,j,m}$ of the $m$th inward link from the neighboring entity
$e_j$ to $e_i$ as follows,
\begin{equation}
\begin{aligned}
\alpha_{i,j,m}=\nu_{\tau}^{T}[ h^{in}_{e_i}|| M_{\tau_m}h^{in}_{e_j}||h_{\tau_m}],\\
\beta_{i,j,m}=\nu_{r}^{T}[ h^{in}_{e_i}|| M_{r_m}h^{in}_{e_j}||h_{r_m} ],
\end{aligned}\label{Equation:weight importance}
\end{equation}
where $||$ denotes the concatenation operator, $\nu_{\tau}^{T}, \nu_{r}^{T}\in\mathbb{R}^{3k}$ are shared temporal and relational attention weight vectors. $h_{e_i}^{in}, h_{e_j}^{in}\in\mathbb{R}^{k}$ are the input features of entities $e_i$ and $e_j$. The entities' input features in the first network layer are their original embeddings. $h_{\tau_m}$ and $h_{r_m}$ are embeddings of the timestamp and relation in the $m$th inward link. Following GAT~\cite{GAT}, we define the normalized element $\omega_{i,j,m}$ and $\upsilon_{i,j,m}$ representing the temporal and relational connectivity from entity $e_i$ to $e_j$ using softmax functions,
\begin{equation}
\begin{aligned}
\omega_{i,j,m}=\frac{\text{exp}(\alpha_{i,j,m})}{\sum_{e_{j}\in\mathcal{N}_{i}^{e}}\sum_{\tau_{m}\in\mathcal{L}^{\tau}_{ij}}\text{exp}(\alpha_{i,j,m})},\\
\upsilon_{i,j,m}=\frac{\text{exp}(\beta_{i,j,m})}{\sum_{e_{j}\in\mathcal{N}_{i}^{e}}\sum_{r_{m}\in\mathcal{L}^{r}_{ij}}\text{exp}(\beta_{i,j,m})},
\end{aligned}\label{Equation:attention weight}
\end{equation}
where $\mathcal{N}^{e}_{i}$ is the set of neighboring entities of $e_i$ and $\mathcal{L}_{ij}^r$ and $\mathcal{L}_{ij}^{\tau}$ denote the sets of relations and time steps in the links from $e_j$ to $e_i$.

The output features $h_{e_i}^{out}$ are obtained with an aggregate which linearly combines the temporal and relational orthogonal transformations of the input features of neighboring entities and a nonlinear ReLU activation function $\sigma(\cdot)$, i.e.,
\begin{equation}
\begin{aligned}
h_{e_i}^{out}=\sigma\Big(&\sum_{{e_{j}\in\mathcal{N}_{i}^{e}}}\sum_{[\tau_{m},r_m]\in\mathcal{L}_{ij}}\omega_{i,j,m}M_{\tau_m}h_{e_j}^{in}\\
&+\upsilon_{i,j,m}M_{r_m}h_{e_j}^{in}\Big).
\end{aligned}\label{Equation:output feature}
\end{equation}
where $\mathcal{L}_{ij}$ denotes the set of links from $e_j$ to $e_i$.

\begin{table*}[t!]
\centering
 \resizebox{0.95\textwidth}{!}{
\begin{tabular}{cccccccccc}
  \toprule
  Dataset & |$\mathcal{E}_1$|& |$\mathcal{E}_2$| &|$\mathcal{R}_1$|&|$\mathcal{R}_2$|&|$\mathcal{T}^*$| &|$\mathcal{Q}_1$|&|$\mathcal{Q}_2$|& |$\mathcal{P}$|& |$\mathcal{S}$|\\
  \midrule
  \textbf{DICEWS-1K/200}&9,517&9,537&247&246&4,017&307,552&307,553&8,566&1,000/200\\
  \textbf{YAGO-WIKI50K-5K/1K}&49,629&49,222&11&30&245&221,050&317,814&49,172&5,000/1,000\\
\textbf{YAGO-WIKI20K}&19,493&19,929&32&130&405&83,583&142,568&19,462&400\\
     \bottomrule

\end{tabular}}

\caption{Statistics of original datasets (not including reverse relations and reverse links).
  }
  \label{tb:dataset}
\end{table*}

\subsection{Entity Alignment Model}
Entity align model aims to embed two KGs into a unified
vector space by pushing the seed alignments of entities together. In this work, the entity align model consists of multiple TEA-GNN layers and a distance function which measures the similarities between final representations of entities.

Let the the $l$-th layer's output features of entity $e_i$ as $h_{e_{i}}^{out(l)}$. A cross-layer representation is employed to capture multi-hoop neighboring information in previous work~\cite{MRAEA} by concatenating output features of different layers. In the same way, we define the global output features $\hat{h}_{e_{i}}^{out}$ of $e_i$ as
\begin{equation}
\begin{aligned}
\hat{h}_{e_{i}}^{out}=[h_{e_{i}}^{out(0)}||h_{e_{i}}^{out(1)}||\cdots||h_{e_{i}}^{out(L)}],
\end{aligned}\label{Equation:final feature}
\end{equation}
where $L$ is the number of layers and ${h}_{e_{i}}^{out(0)}={h}_{e_{i}}^{in}$ are the input features.

We further concatenate the average embeddings of connected timestamps with output features of entities to get multi-view embeddings as final entity representations, i.e.,
%. The multi-view entity representations
\begin{equation}
\begin{aligned}
h^{mul}_{e_i}=[ \hat{h}_{e_{i}}^{out}|| \frac{1}{|\mathcal{N}_{e_i}^{\tau}|}\sum_{\tau_{m}\in\mathcal{N}_{e_i}^{\tau}}h_{\tau_m}],
\end{aligned}\label{Equation:multi-view}
\end{equation}
where $\mathcal{N}_{e_i}^{\tau}$ represents the set of timestamps around entity $e_i$.

Entity alignments are predicted based on the distances between the final output features of entities from two KGs. For two entities $e_i\in\mathcal{E}_1$ and $e_j\in\mathcal{E}_2$ from different sources, we use L1 distance to measure the distance between them as follows,
\begin{equation}
\begin{aligned}
d(e_i,e_j)=||{h}_{e_{i}}^{mul}-{h}_{e_{j}}^{mul}||_{_1},
\end{aligned}\label{Equation:distance}
\end{equation}

A margin rank loss is used as the optimization objective of the entity align model, i.e.,
\begin{equation}
\begin{aligned}
&\mathcal{L}=\sum_{(e_i,e_j)\in\mathcal{S}}\sum_{(e_i,e'_j)\in\mathcal{S}'}\sigma(d(e_i,e_j)+\lambda-d(e_i,e'_j))\\
&+\sum_{(e_i,e_j)\in\mathcal{S}}\sum_{(e'_i,e_j)\in\mathcal{S}'}\sigma(d(e_i,e_j)+\lambda-d(e'_i,e_j)).
\end{aligned}\label{Equation:loss}
\end{equation}
where $\lambda$ denotes the margin, $\mathcal{S}'$ is the set of generated negative entity pairs, $e'_i\in\mathcal{E}_1$ and $e'_j\in\mathcal{E}_2$ are the negative entities of $e_i$ and $e_j$.
Negative entities are sampled randomly and an RMSprop optimizer is used to minimize the loss function. 

During testing, we adopt CSLS~\cite{CSLS} as the distance metric to measure similarities between entity embeddings.

\section{Experiments}
\subsection{Datasets}~\label{sec:datasets}
In this work, we build five datasets from ICEWS05-15~\cite{TA-TransE}, YAGO3~\cite{YAGO3} and Wikidata~\cite{Wikidata} to evaluate our approach. 

ICEWS05-15\footnote{https://github.com/nle-ml/mmkb} is originally extracted from ICEWS~\cite{ICEWS} which is a repository that contains political events with specific time annotations, e.g.\ (\text{Barack Obama}, \text{Make a visit}, \text{Ukraine}, \text{2014-07-08}). It is noteworthy that time annotations in ICEWS are all time points. ICEWS05-15 contains events during 2005 to 2015. We build two datasets \textbf{DICEWS-1K} and \textbf{DICEWS-200} in the similar way to the construction of DFB datasets~\cite{IPTransE}. We first randomly divide ICEWS05-15 quadruples into two subsets $\mathcal{Q}_1$ and $\mathcal{Q}_2$ of similar size, and make the overlap ratio of the amount of shared quadruples between $\mathcal{Q}_1$ and $\mathcal{Q}_2$ to all quadruples equal to 50\%. The only difference between DICEWS-1K and DICEWS-200 is the number of alignment seed $\mathcal{S}$. In DICEWS-1K and DICEWS-200, i.e., 1,000 and 200 of entity pairs between TKGs are pre-known. The time unit of ICEWS datasets is 1 day, which means that each day is an individual time step. 

\begin{table*}[t!]
\centering
\resizebox{1\textwidth}{!}{
\begin{tabular}{ccccccccccccc}
    \toprule
\multirow{2}{*}{Models} &\multicolumn{3}{c}{\textbf{DICEWS-1K}}&\multicolumn{3}{c}{\textbf{DICEWS-200}}&\multicolumn{3}{c}{\textbf{YAGO-WIKI50K-5K}}&\multicolumn{3}{c}{\textbf{YAGO-WIKI50K-1K}}\cr 
 \cmidrule(lr){2-4}\cmidrule(lr){5-7}\cmidrule(lr){8-10}\cmidrule(lr){11-13}
&MRR&Hits@1&Hits@10&MRR&Hits@1&Hits@10&MRR&Hits@1&Hits@10&MRR&Hits@1&Hits@10\cr
\midrule
MTransE&.150&.101&.241&.104&.067&.175&.322&.242&.477&.033&.012&.067\cr
JAPE&.198&.144&.298&.138&.098&.210&.345&.271&.488&.157&.101&.262\cr
AlignE&.593&.508&.751&.303&.222&.457&.800&.756&.883&.618&.565&.714\cr
\midrule
GCN-Align&.291&.204&.466&.231&.165&.363&.581&.512&.711&.279&.217&.398\cr
MRAEA&.745&.675&.870&.564&.476&.733&.848&.806&.913&.685&.623&.801\cr
RREA&.780&.722&.883&.719&.659&.824&.868&.828&.938&.753&.696&.859\cr
\midrule
TU-GNN&.693&.610&.848&.610&.518&.788&.839&.795&.916&.712&.647&.834\cr
TEA-GNN&\textbf{.911}&\textbf{.887}&\textbf{.947}&\textbf{.902}&\textbf{.876}&\textbf{.941}&\textbf{.909}&\textbf{.879}&\textbf{.961}&\textbf{.775}&\textbf{.723}&\textbf{.871}\cr
\bottomrule

\hline
\end{tabular}}
\caption{
Entity alignment results on ICEWS and YAGO-WIKI50K datasets. The best results are written bold.
    }
    \label{tb: main results}
\end{table*}

YAGO3 and Wikidata are two common large-scale knowledge bases containing time information of various forms including time points, beginning or end time, and time intervals. Lacroix et al.~\shortcite{TComplEx} extract a subset\footnote{https://github.com/facebookresearch/tkbc} from Wikidata in which 90\% of facts are non-temporal while others have time annotations attached. We select top 50,000 entities according to their frequencies in Wikidata and link them to their equivalent YAGO entities\footnote{http://resources.mpi-inf.mpg.de/yago-naga/yago3.1} according to their QIDs and the mappings of YAGO entities to Wikidata QIDs. We generate two TKGs only involving the selected entities from the original Wikidata dataset and all YAGO facts, and then attach complementary time information
%\footnote{http://resources.mpi-inf.mpg.de/yago-naga/yago3.1/yagoMetaFacts.ttl.7z}
to meta YAGO facts. We build two time-aware datasets \textbf{YAGO-WIKI50K-5K} and \textbf{YAGO-WIKI50K-1K} by removing non-temporal facts in the generated TKGs and using different numbers of alignment seeds $\mathcal{S}$. In addition, we build a hybrid dataset \textbf{YAGO-WIKI20K} containing
both temporal and non-temporal facts with 400 pairs of alignment seeds by reducing sizes of entity sets of two TKGs to around 20,000. To generate the shared time set $\mathcal{T}^*$ for a YAGO-WIKI dataset, we drop month and date information and use the first time step $\tau_0$ to represent unobtainable time information.

Statics of all datasets are listed in Table 1. $\mathcal{P}$ denotes the set of reference entity pairs. The set of reference entity pairs other than pre-aligned entity pairs, i.e., $\mathcal{P}-\mathcal{S}$ are used for testing.
\subsection{Experimental Setup}~\label{sec: setup}
Following the previous work, we perform entity alignment as a ranking task based on similarities between entity embeddings, and use Mean Reciprocal Rank (MRR) and Hits@N (N=1, 10) as evaluation metrics. The default configuration of our model is as follows: embedding dimension $k=100$, learning rate $lr=0.005$, number of TEA-GNN layers $L=2$, margin $\gamma=1$ and dropout rate is 0.3. Below we only list the non-default hyperparamters: $\gamma=3$ for DICEWS-200 and YAGO-WIKI20K; $k=25$ for YAGO-WIKI50K-5K and YAGO-WIKI50K-1K. To verify the effectiveness of integration of time information, we implement a time-unaware variant of TEA-GNN which takes all time steps $\tau_i\in\mathcal{T}^*$ as unknown time information $\tau_0$, denoted as TU-GNN. The non-default hyperparameters of TU-GNN are as follows: $\gamma=3$ for DICEWS-1K and YAGO-WIKI20K; $\gamma=5$ for DICEWS-200; $k=25$ for YAGO-WIKI50K datasets. The reported performance is the average of five
independent training runs.

In this work, we compare our proposed models with three strong translational baseline models and three state-of-the-art GNN-based models including MTransE~\cite{MTransE}, JAPE~\cite{JAPE}, AlignE~\cite{BootEA}, GCN-Align~\cite{GCN-Align}, MRAEA~\cite{MRAEA} and RREA~\cite{RREA}. We choose AlignE instead of BootEA since we do not use iterative learning for other models including our proposed models. Due to the lack of attribute information, we use the SE (Structural Embedding) variants of JAPE and GCN-Align as baseline models. Except that the experiments of MTransE is implemented based on OpenEA framework~\cite{OpenEA}, all experiments of baseline models are implemented based on their resource codes. All target models including our proposed models are trained on a GeForce GTX 1080Ti GPU. For a fair comparison, we set the maximum embedding dimension as 100 for all target models. Details of implementation and grid research for hyperparameters can be found in Appendix~\ref{sec:Implementation Details}.

\begin{table*}[t!]
\centering
\resizebox{1\textwidth}{!}{
\begin{tabular}{ccc}
    \toprule
Entities to Be Aligned& Predictions & Similar Facts (Links) Involving Aligned Entities Between $\mathcal{Q}_{1}$ and $\mathcal{Q}_{2}$  \cr
\midrule
\multirowcell{6}{Daniel Scioli\\ (in $\mathcal{E}_1$ of DICEWS-200)}&\multirowcell{6}{TEA-GNN:\\ Daniel Scioli\\ \\TU-GNN: \\ Agustín Rossi} & (Daniel Scioli, Accuse, Senate (Argentina), 2015-06-28), \cr
& & (Presidential Candidate (Argentina), Make Statement,  Daniel Scioli, 2015-03-07), \cr
& & $\dots$ (in $\mathcal{Q}_1$ of DICEWS-200)\cr
& & (Agustín Rossi, Accuse, Senate (Argentina), 2009-08-26), \cr
& & (Presidential Candidate (Argentina), Make Statement,  Agustín Rossi, 2015-04-18),\cr
& & $\dots$ (in $\mathcal{Q}_2$ of DICEWS-200)\cr
\midrule
\multirowcell{4}{Leon Benko (Q1389599)\\ (in $\mathcal{E}_2$ of YAGO-WIKI50K-1K)}&\multirowcell{4}{TEA-GNN:\\ <Leon\_Benko> \\TU-GNN: \\ <Olivier\_Fontenette>} & (<Olivier\_Fontenette>, <playsFor>, <K.V.\_Kortrijk>, [2008, -]), \cr
& & $\dots$ (in $\mathcal{Q}_1$ of YAGO-WIKI50K-1K)\cr
& & (Leon Benko (Q1389599), member of sports team,  K.V. Kortrijk (Q618620), [2009, 2010]),\cr
& & $\dots$ (in $\mathcal{Q}_2$ of YAGO-WIKI50K-1K)\cr
\bottomrule
\end{tabular}}
\caption{
Examples of Different Alignment Predictions between TEA-GNN and TU-GNN.
    }
    \label{tb: case analysis}
\end{table*}

\begin{figure*}[t!]
\centering
\includegraphics[width=15cm]{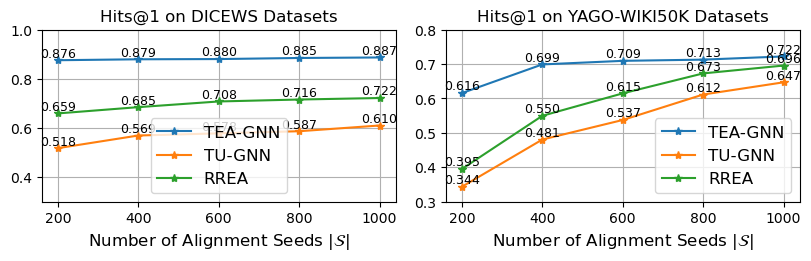} 
\caption{Hits@1 of TEA-GNN, TU-GNN and RREA on entity alignment, w.r.t. number of alignment seeds |$\mathcal{S}$|.}
\label{fig:seed}
\end{figure*}

\subsection{Results and Analysis}
\paragraph{Main Results}
Table~\ref{tb: main results} shows the entity alignment results of our proposed models and all baselines on ICEWS and YAGO-WIKI50K datasets. It can be shown that TEA-GNN remarkably outperforms all baseline models on four TKG datasets across all metrics. Compared to RREA which achieves the best results among than all baseline models, TEA-GNN obtains the improvement of 22.9\%, 32.9\%, 6.2\% and 3.9\% regarding Hits@1 on four TKG datasets, respectively.

\paragraph{Qualitative Study}
To study the effect of the integration of time information on the entity alignment performances of TEA-GNN, we conduct a qualitative study of TEA-GNN and its time-unaware variant TU-GNN. Table~\ref{tb: case analysis} lists several examples that TEA-GNN gives different predictions from TU-GNN with consideration of additional time information. In the first case, TU-GNN wrongly aligns two entities from $\mathcal{G}_1$ and $\mathcal{G}_2$ of DICEWS-200, i.e., Daniel Scioli and Agustín Rossi, because these two entities have very similar connected links in $\mathcal{G}_1$ and $\mathcal{G}_2$ regardless of time information. As shown in Table~\ref{tb: case analysis}, some links respective to these two entities in $\mathcal{G}_1$ and $\mathcal{G}_2$ have the same linked entities and relation types, leading to the result that TU-GNN identifies them as an equivalent entity pair. On the other hand, TEA-GNN can correctly distinguish these two entities since the relevant links have different timestamps. Similarly, TU-GNN recognizes a Wikidata entity Leon Benko (Q1389599) and a YAGO entity <Olivier\_Fontenette> as the same person since these two person played for the same football club, while TEA-GNN can learn that they played for different periods and thus are not the same person in real world. These cases demonstrate the effect of time information on the performances of our proposed entity alignment models.

\begin{table*}[t]
\centering
\resizebox{0.9\textwidth}{!}{
\begin{tabular}{cccccccccc}
    \toprule
\multirow{2}{*}{} &\multicolumn{3}{c}{\textbf{Highly Time-Sensitive}}&\multicolumn{3}{c}{\textbf{Lowly Time-Sensitive}}&\multicolumn{3}{c}{\textbf{In Total}}\cr 
 \cmidrule(lr){2-4}\cmidrule(lr){5-7}\cmidrule(lr){8-10}
&MRR&Hits@1&Hits@10&MRR&Hits@1&Hits@10&MRR&Hits@1&Hits@10\cr
\midrule
TEA-GNN&.888&853&.950&.364&.319&.449&.553&.512&.630\cr
TU-GNN&.790&.737&.888&.366&.315&.463&.519&.468&.617\cr
\bottomrule

\hline
\end{tabular}}
\caption{
Entity alignment results on different test sets of YAGO-WIKI20K.
    }
    \label{tb: part results}
\end{table*}

\begin{figure*}[t!]
\centering
\includegraphics[width=15.5cm]{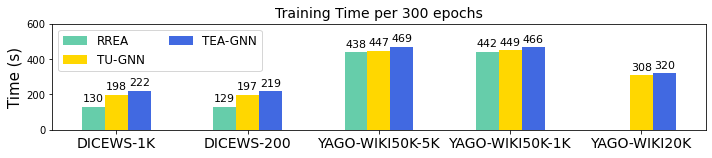} 
\caption{Training time per 300 epochs on different datasets.}
\label{fig:training time}
\end{figure*}

\paragraph{Sensitivity Study} 
In this study, we would like to answer the following research questions: 1. Does the time-aware EA method have a better robustness to the size of pre-aligned entity pairs than time-agnostic ones?  2. In our method, what effect does the inclusion of time information have on entities with different time sensitivities?

Numerically, compared to TU-GNN, TEA-GNN improves Hits@1 by 45.4\% and 69.1\% on DICEWS-1K and DICEWS-200, 10.6\% and 11.6\% on YAGO-WIKI50K-5K and YAGO-WIKI50K-1K. 
It can be shown that the improvements on datasets with less alignment seeds are more significant. To futher verify this observation, we evaluate the performances of these two models and RREA on ICEWS and YAGO-WIKI50K datasets with different numbers of alignment seeds. As shown in Figure~\ref{fig:seed}, the performance difference between TEA-GNN and two time-unaware models becomes greater with the decreasing of the numbers $|\mathcal{S}|$ of alignment seeds from 1000 to 200. In practical applications, alignment seeds are difficult to obtain. Since our method performs well with a small amount of pre-aligned entity pairs, it can more easily be applied in large-scale KGs compared to time-unaware EA methods.

We also conduct a study on the prediction accuracy of aligned entities which have different time sensitivity. As mentioned in Section~\ref{sec:datasets}, we generate a hybrid dataset YAGO-WIKI20K where 17.5\% of YAGO facts and 36.6\% of Wikidata facts are non-temporal. We divide all testing entity pairs in this dataset into two categories based on their sensitivity to time information, i.e., \textbf{highly time-sensitive} entity pairs and \textbf{lowly time-sensitive} entity pairs. Time sensitivity $s_i$ of a single entity $e_{i}$ is defined as the ratio of the number of its time-aware connected links in which $\tau \neq \tau_0$ over the total number of all links $\mathcal{L}_{i}$ within its neighborhood, i.e.,

\begin{equation}
s_{i}=1-{|\mathcal{L}^{\tau_0}_{i}|}/{|\mathcal{L}_{i}|},
\end{equation}\label{Equation:Sens}
where $\mathcal{L}^{\tau_0}_{i}$ denotes the set of time-unaware links connecting $e_{i}$. Given an entity pair $(e_{i1},e_{i2})$ between $\mathcal{G}_1$ and $\mathcal{G}_2$, we call them as a highly time-sensitive entity pair if $s_{i1}\geqslant 0.5$ and $s_{i2}\geqslant 0.5$. Otherwise, they are lowly time-sensitive. 

Among 19,062 testing entity pairs of YAGO-WIKI20K, 6,898 of them are highly time-sensitive and others are lowly time-sensitive according to the above definitions. The entity alignment results of TEA-GNN and TU-GNN on the highly time-sensitive test set and the lowly time-sensitive test set are reported in Table 4. It can be shown that TEA-GNN and TU-GNN have close performance on entity alignment for lowly time-sensitive entity pairs while TEA-GNN remarkably outperforms TU-GNN on the highly time-sensitive test set. In other words, the effect of incorporation of time information are more significant when testing entity pairs are more time-sensitive. 

\paragraph{Complexity Study} Given two TKGs to be aligned, i.e., $\mathcal{G}_1=(\mathcal{E}_1,\mathcal{R}_1,\mathcal{T}^{*},\mathcal{Q}_1)$ and $\mathcal{G}_2=(\mathcal{E}_2,\mathcal{R}_2,\mathcal{T}^{*},\mathcal{Q}_2)$, the total number of trainable parameters $|p|$ of TEA-GNN is equal to,
\begin{equation}
\begin{aligned}
|p|=&k\times(|\mathcal{E}_1|+|\mathcal{E}_2|+2|\mathcal{R}_1|+2|\mathcal{R}_2|\\&+|\mathcal{T}^{*}|)
+3k\times L+3k\times L, 
\end{aligned}
\end{equation}\label{Equation:para num}
where $L$ denotes the number of TEA-GNN layers, the last two terms represent numbers of parameters of shared temporal and relational attention weight vectors involved in Equation~\ref{Equation:weight importance}. Compared to parameter-efficient translational entity align models like MTransE, JAPE, in which the numbers of parameters are $k\times(|\mathcal{E}_1|+|\mathcal{E}_2|+|\mathcal{R}_1|+|\mathcal{R}_2|)$, TEA-GNN uses additional parameters only for reverse relation embeddings, time embeddings and attention weight vectors, which are much fewer than parameters
of entity embeddings in most cases.

As shown in Figure~\ref{fig:training time}, the processing of the additional time information does not excessively increase the training time for TEA-GNN, compared to RREA and TU-GNN. Since we set the maximum number of epochs as 6000, the training processes of our proposed models on different datasets can be completed within a couple of hours on a single GeForce GTX 1080Ti GPU.

\section{Conclusion}
The main contributions of this paper are threefold:
\begin{itemize}[leftmargin=*]
\setlength{\topsep}{0pt}
\setlength{\itemsep}{3pt}
\setlength{\parsep}{1pt}
\setlength{\parskip}{0pt}
    \item We propose a novel GNN-based approach TEA-GNN which can model temporal relational graphs with an orthogonal transformation based time-aware attention mechanism and perform entity alignment tasks between TKGs. To the best of our knowledge, this work is the first attempt to integrate time information into an embedding-based entity alignment approach.
    \item Existing temporal GNN models typically discretize a temporal graphs into multiple static snapshots and utilize a combination of GNNs and recurrent architectures. Differently, we treat timestamps as attentive properties of links between nodes. This method has been proven to be time-efficient in our case and could potentially be used for non-relational temporal graph representation learning.
    \item Multiple new datasets are created in this work for evaluating the performance of entity alignment models on TKGs. Experiments show that TEA-GNN remarkably outperforms the state-of-the-art entity alignment models on various well-built TKG datasets. 
\end{itemize}

For future work, we will try to integrate other types of information, e.g,. attribute information, into our model and extend our model for other learning tasks of temporal graphs.

\section*{Acknowledgements}
We acknowledge the support of the following projects: SPEAKER (BMWi FKZ 01MK20011A), JOSEPH (Fraunhofer Zukunftsstiftung), the EU projects Cleopatra (GA 812997), PLATOON(GA 872592), TAILOR(EU GA 952215), the BMBF projects MLwin(01IS18050) and the BMBF excellence clusters ML2R (BmBF FKZ 01 15 18038 A/B/C), ScaDS.AI (IS18026A-F) and the China Scholarship Council (CSC).
% \section*{Acknowledgements}
% This work is supported by the CLEOPATRA project (GA no.~812997), the German national funded BmBF project MLwin, the BOOST project and the China Scholarship Council (CSC).

\bibliography{anthology,custom}
\bibliographystyle{acl_natbib}

\clearpage
\appendix

\section{Implementation Details}~\label{sec:Implementation Details}
As mentioned in Section~\ref{sec: setup}, we use the source codes~\footnote{https://github.com/nju-websoft/JAPE}\footnote{https://github.com/nju-websoft/BootEA}\footnote{https://github.com/1049451037/GCN-Align/}\footnote{https://github.com/MaoXinn/MRAEA}\footnote{https://github.com/MaoXinn/RREA} respective to baseline models for evaluation, except that we evaluate MTransE based on the implementation of OpenEA framework~\footnote{https://github.com/nju-websoft/OpenEA/}.

For all baseline models, we mostly follow their default optimal configurations regarding learning rates $lr$, batch sizes $b$, negative sampling rates $\eta$, dropout rates $dr$, numbers of GNN layers $L$ and mainly focus on the grid research of embedding dimensions $k$ and margins $\gamma$ (negative weights $\alpha$ for JAPE ). We also follow the respective original papers to fix the numbers of multi-head attention mechanisms as 2 for MRAEA and set the balance weight $\beta=0.9$ for GCN-Align. For all baseline models and our proposed models, we tune $k$ in the range of $(25,50,75,100)$, and $\gamma$ or $\alpha$ in the range of $(0,0.5,1,2,3,5,7,10,15,20)$. Specially, we use the same margin hyperparameters as the original paper for AlignE. To make a fair comparison, we use the same setup for our proposed model as MRAEA and RREA to fix $L=2$, $dr=0.3$, $b=|\mathcal{E}_1|+|\mathcal{E}_2|$ and $\eta = b//|\mathcal{S}|+1$ where $//$ denotes the round-down after division, and also conduct the same grid research of embedding dimensions $k$ and margins $\gamma$ for TEA-GNN and TU-GNN as what we do for baseline models. 

All hyperparameters used for getting the results reported in Table~\ref{tb: main results} and~\ref{tb: part results} are listed in Table~\ref{tb:hp1}, \ref{tb:hp2}, \ref{tb:hp3}, \ref{tb:hp4} and~\ref{tb:hp5}.
In our experiments, for MRAEA, RREA and our proposed models, the usage of the higher-dimensional embeddings
would lead to out-of-memory problems since we only use a single mid-range GPU device for training and the learning of large-scale graph neural networks with numerous nodes and high-dimensional embeddings needs excessive memory footprint during the training process. It is predictable that we can possibly further boost the performances of TU-GNN and TEA-GNN on YAGOWIKI50K datasets by increasing their embedding dimensions with more high-performance GPU devices.

The source codes and datasets used in this work are submitted as the supplementary materials for reproducibility and will be released on Github after the anonymity period.

\begin{table}[h]
\centering
 \resizebox{0.49\textwidth}{!}{
\begin{tabular}{ccccccc}
  \toprule
  Models & $k$ & $lr$ & $b$ &$\eta$ &$\gamma$ (or $\alpha$)&$dr$ \\
  \midrule
  MTransE&100&0.01&20,000&10&10&-\\
  JAPE&100&0.01&10,000&1&2&-\\
  AlignE&100&0.01&20,000&10&0.01,2,0.7&-\\
  GCN-Align&100&20&-&5&3&0\\
  MRAEA&100&0.001&19,054&20&1&0.3\\
  RREA&100&0.005&19,054&20&3&0.3\\
  TU-GNN&100&0.005&19,054&20&3&0.3\\
  TEA-GNN&100&0.005&19,054&20&1&0.3\\
  \bottomrule

\end{tabular}}
\vspace{-0.1cm}
\caption{Hyperparameters of target models for DICEWS-1K.
  }
  \vspace{-0.24cm}
  \label{tb:hp1}
\end{table}

\begin{table}[h]
\centering
 \resizebox{0.49\textwidth}{!}{
\begin{tabular}{ccccccc}
  \toprule
  Models & $k$ & $lr$ & $b$ &$\eta$ &$\gamma$ (or $\alpha$)&$dr$ \\
  \midrule
  MTransE&100&0.01&20,000&10&7&-\\
  JAPE&100&0.01&10,000&1&3&-\\
  AlignE&100&0.01&20,000&10&0.01,2,0.7&-\\
  GCN-Align&100&20&-&5&7&0\\
  MRAEA&100&0.001&19,054&96&2&0.3\\
  RREA&100&0.005&19,054&96&2&0.3\\
  TU-GNN&100&0.005&19,054&96&5&0.3\\
  TEA-GNN&100&0.005&19,054&96&3&0.3\\
  \bottomrule

\end{tabular}}
\vspace{-0.1cm}
\caption{Hyperparameters of target models for DICEWS-200.
  }
  \vspace{-0.24cm}
  \label{tb:hp2}
\end{table}

\begin{table}[h]
\centering
 \resizebox{0.49\textwidth}{!}{
\begin{tabular}{ccccccc}
  \toprule
  Models & $k$ & $lr$ & $b$ &$\eta$ &$\gamma$ (or $\alpha$)&$dr$ \\
  \midrule
  MTransE&100&0.01&20,000&10&10&-\\
  JAPE&100&0.01&10,000&1&3&-\\
  AlignE&100&0.01&20,000&10&0.01,2,0.7&-\\
  GCN-Align&100&20&-&5&3&0\\
  MRAEA&75&0.001&98,851&20&1&0.3\\
  RREA&50&0.005&98,851&20&1&0.3\\
  TU-GNN&25&0.005&98,851&20&1&0.3\\
  TEA-GNN&25&0.005&98,851&20&1&0.3\\
  \bottomrule

\end{tabular}}
\vspace{-0.1cm}
\caption{Hyperparameters of target models for YAGO-WIKI50K-5K.
  }
  \vspace{-0.24cm}
  \label{tb:hp3}
\end{table}

\begin{table}[h]
\centering
 \resizebox{0.49\textwidth}{!}{
\begin{tabular}{ccccccc}
  \toprule
  Models & $k$ & $lr$ & $b$ &$\eta$ &$\gamma$ (or $\alpha$)&$dr$ \\
  \midrule
  MTransE&100&0.01&20,000&10&1&-\\
  JAPE&100&0.01&10,000&1&2&-\\
  AlignE&100&0.01&20,000&10&0.01,2,0.7&-\\
  GCN-Align&100&20&-&5&10&0\\
  MRAEA&75&0.001&98,851&99&2&0.3\\
  RREA&50&0.005&98,851&99&1&0.3\\
  TU-GNN&25&0.005&98,851&99&1&0.3\\
  TEA-GNN&25&0.005&98,851&99&1&0.3\\
  \bottomrule

\end{tabular}}
\vspace{-0.1cm}
\caption{Hyperparameters of target models for YAGO-WIKI50K-1K.
  }
  \vspace{-0.24cm}
  \label{tb:hp4}
\end{table}

\begin{table}[h]
\centering
 \resizebox{0.49\textwidth}{!}{
\begin{tabular}{ccccccc}
  \toprule
  Models & $k$ & $lr$ & $b$ &$\eta$ &$\gamma$ (or $\alpha$)&$dr$ \\
  \midrule
  TU-GNN&100&0.005&39,422&99&3&0.3\\
  TEA-GNN&100&0.005&39,422&99&3&0.3\\
  \bottomrule

\end{tabular}}
\vspace{-0.1cm}
\caption{Hyperparameters of target models for YAGO-WIKI20K.
  }
  \vspace{-0.24cm}
  \label{tb:hp5}
\end{table}

\clearpage

% \section{Space Complexity}
% \label{sec:Space Complexity}
% Given two TKGs to be aligned, i.e., $\mathcal{G}_1=(\mathcal{E}_1,\mathcal{R}_1,\mathcal{T}^{*},\mathcal{Q}_1)$ and $\mathcal{G}_2=(\mathcal{E}_2,\mathcal{R}_2,\mathcal{T}^{*},\mathcal{Q}_2)$, numbers of parameters of different entity alignment models are listed in Table.

% \begin{table}[h]
% \centering
%  \resizebox{0.49\textwidth}{!}{
% \begin{tabular}{cc}
%   \toprule
%   Models & Parameters\\
%   \midrule
%   \multirowcell{2}{RREA}&\multirowcell{2}{$k\times(|\mathcal{E}_1|+|\mathcal{E}_2|+2|\mathcal{R}_1|+2|\mathcal{R}_2|)$\\  $+3k\times L$} \cr\cr
%   \midrule
% \multirowcell{2}{TU-GNN}&\multirowcell{2}{$k\times(|\mathcal{E}_1|+|\mathcal{E}_2|+2|\mathcal{R}_1|+2|\mathcal{R}_2|+1)$\\  $+3k\times L+3k\times L$} \cr\cr
%   \midrule
%   \multirowcell{2}{TEA-GNN}&\multirowcell{2}{$k\times(|\mathcal{E}_1|+|\mathcal{E}_2|+2|\mathcal{R}_1|+2|\mathcal{R}_2|+|\mathcal{T}^{*}|)$\\  $+3k\times L+3k\times L$} \cr\cr

%      \bottomrule

% \end{tabular}}
% \vspace{-0.1cm}
% \caption{Statistics of original datasets (not including reverse relations and reverse links).
%   }
%   \vspace{-0.24cm}
%   \label{tb:}
% \end{table}

\end{document}